\pdfoutput=1
\documentclass{article}
\usepackage{spconf,amsmath,graphicx}

\usepackage{cite}
\usepackage{amsmath,amssymb,amsfonts}
\usepackage{algorithmic}
\usepackage{graphicx}
\usepackage{textcomp}
\usepackage[table]{xcolor}
\usepackage{booktabs}
\usepackage{multirow}
\usepackage{adjustbox}

\definecolor{green}{rgb}{0.12, 0.85, 0.6} 

\definecolor{red}{rgb}{0.89, 0.09, 0.29} 
\newcommand{\perplus}[1]{{\color{red}{\scriptsize\bf (+#1)}}}
\definecolor{blue}{rgb}{0.316, 0.818, 0.996}

\usepackage{colortbl}
\definecolor{defaultcolor}{gray}{.9}

\title{AdaViPro: Region-based Adaptive Visual Prompt \\for Large-Scale Models Adapting}
\name{}
\address{}
\name{Mengyu Yang\textsuperscript{1}, 
Ye Tian\thanks{* Corresponding author. This work was supported in part by the National Natural Science Foundation of China under Grant 62072048, in part by Industry-University-Research Innovation Fund of Universities in China under Grant 2021ITA07005, and in part by BUPT Excellent Ph. D. Students Foundation under Grant CX20242004. }\textsuperscript{1}*,
Lanshan Zhang\textsuperscript{1}, 
Xiao Liang\textsuperscript{2}, 
Xuming Ran\textsuperscript{3}, 
Wendong Wang\textsuperscript{1}}
\address{
\textsuperscript{1}
Beijing University of Posts and Telecommunications, Beijing, China,\\
\textsuperscript{2}Xidian University, Xian, China, \\
\textsuperscript{3}Dalian University of Technology, Dalian, China
}

\begin{document}
\topmargin=0mm
%
\maketitle
\begin{abstract}
Recently, prompt-based methods have emerged as a new alternative `parameter-efficient fine-tuning' paradigm, which only fine-tunes a small number of additional parameters while keeping the original model frozen.
However, despite achieving notable results, existing prompt methods mainly focus on `what to add', while overlooking the equally important aspect of `where to add', typically relying on the manually crafted placement.
To this end, we propose a region-based \textit{\textbf{Ada}ptive \textbf{Vi}sual \textbf{Pro}mpt}, named AdaViPro, which integrates the `where to add' optimization of the prompt into the learning process.
Specifically, we reconceptualize the `where to add' optimization as a problem of regional decision-making.
During inference, AdaViPro generates a regionalized mask map for the whole image, which is composed of 0 and 1, to designate whether to apply or discard the prompt in each specific area.
Therefore, we employ Gumbel-Softmax sampling to enable AdaViPro's end-to-end learning through standard back-propagation.
Extensive experiments demonstrate that our AdaViPro yields new efficiency and accuracy trade-offs for adapting pre-trained models.
\end{abstract}

\section{Introduction}
Recently, fuelled by the label-free self-supervised learning, Transformer \cite{transformer} and its variants have demonstrated remarkable success in a wide range of areas, such as natural language processing \cite{bert,gpt3}, computer vision \cite{wang2023improving,vimo} and audio signal processing \cite{liu2023fine}.
However, as the scale of the model size grows, the conventional pretraining-finetuning paradigm becomes impractical \cite{dam-vp, actionclip}.
Fully fine-tuning these large pre-trained models for varied downstream tasks proves to be expensive and inflexible.
Meanwhile, due to the limited training data of downstream tasks, improper fine-tuning might compromise the universal representations established during pre-training, diminishing their effectiveness.

\begin{figure}[t]
\centerline{\includegraphics[width=0.45\textwidth]{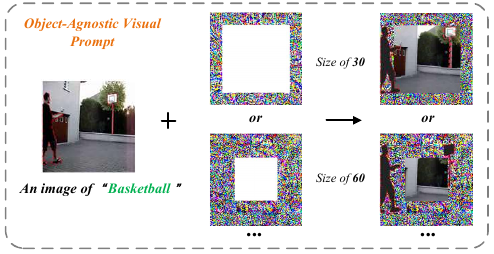}}
    \caption
    { Visualization of the pixel-level VP \cite{vp} of 30 and 60 prompt widths.
    Available information about the label is covered by the fixed-position prompt.   
    }
    \label{fig1}
    \vspace{-0.2 in}
\end{figure}

To break this dilemma, prompt-based fine-tuning methods \cite{coop,vp,vpt,ju2022prompting,actionclip} introduce an alternative paradigm to adapt the large-scale pre-trained models. 
Originally developed in NLP \cite{prompt}, prompt-tuning preserves the parameters in a frozen state throughout the fine-tuning process. 
It achieves alignment between the pre-training initialization and the downstream task by strategically modifying the input format with task-relevant descriptions.
Different tasks adopt distinct descriptions to bootstrap the model for task-specific adaptation.
In this new paradigm, various downstream tasks are allowed to utilize the same pre-trained parameters, thereby fostering the parameter-efficient fine-tuning process.
Following the success of NLP, this elegantly simple but effective paradigm has also been brought to computer vision \cite{vpt,vp}.
Some recent works propose to append learnable tokens to the tokenized sequence or superimpose extra learnable noises onto the input image.
The former format \cite{vpt}, modifying the token sequence, closely resembles prefix-tuning in NLP, where the learnable tokens are fed into the model as additional elements along with the original sequence.
While the latter format \cite{vp} directly modifies the image with task-specific `mosaics' at fixed positions on a pixel-level.
With these input modifications, both formats have efficiently adapted pre-trained models to downstream tasks, while substantially reducing tuning parameters.

\begin{figure}[t]
\centerline{\includegraphics[width=0.5\textwidth]{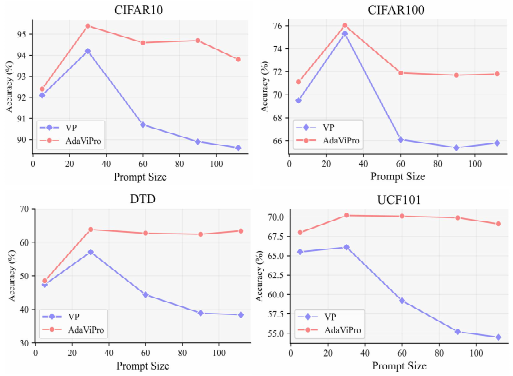}}
    \caption
    { 
    Performance of VP \cite{vp} and AdaViPro on CIFAR10, CIFAR100, DTD and UCF101 across various prompt sizes of \{5, 60, 90, 112\}.
    }
    \label{fig2}
    \vspace{-0.2 in}
\end{figure}

The core issues in prompt engineering should revolve around two aspects: `what to add' and `where to add'.
Despite achieving notable results, existing methods predominantly focus on `what to add', while overlooking the equally important aspect of `where to add'.
To optimize `what to add', prompts are abstracted as learnable parameters, which participate in end-to-end training.
However, the placement of them is hand-crafted, making it challenging to achieve optimal effectiveness.
As shown in Figure \ref{fig1}(a), the visual prompt is a double-edged sword, providing auxiliary information while drowning out some of the raw information.
When the prompt size reaches 60, there is almost a complete absence of available information about the label.
In Figure \ref{fig2}, it is also evident that with the increase in prompt size, the VP's performance undergoes a steep decline.
Therefore, due to the size and distribution of the objects being variable, it is clear that the fixed-position prompt cannot adapt to such variability, even disturbing the judgment of the model.

With this in mind, we propose a region-based \textbf{Ada}ptive \textbf{Vi}sual \textbf{Pro}mpt method, named AdaViPro, for efficiently adapting large-scale pre-trained models in downstream tasks. 
In contrast to the existing hand-crafted prompt position, AdaViPro incorporates the optimization of `where to add' into the learning process.
Specifically, we transform the question of `where to add' into a regional decision-making problem.
AdaViPro generates a regionalized mask map for the whole image, which is composed of 0 and 1, to indicate whether to apply or discard the prompt in their area, respectively.
During training, since generating a mask map composed of $\{0,1\}$ is discrete and non-differentiable, we utilize Gumbel-Softmax sampling to facilitate the learning of mask generator through standard back-propagation.
Extensive experiments are conducted with various prompt sizes on nine image benchmarks, showing that AdaViPro significantly enhances the transfer effectiveness across all the prompt sizes in adapting pre-trained models. It can be seen in Figure \ref{fig2}.

The major contributions are summarized as follows:
\begin{itemize}
    \item We propose AdaViPro, a region-based adaptive visual prompt method, which generates image-specific prompts, contrasting with existing fixed-position prompts.
    \item We utilize Gumbel-Softmax sampling to facilitate the end-to-end learning of the whole framework with standard back-propagation.
    \item Extensive experiments on nine image benchmarks demonstrate that our method significantly enhances the effectiveness of adapting pre-trained models.
\end{itemize}

\begin{figure*}[t]
    \centerline{\includegraphics[width=0.95\textwidth]{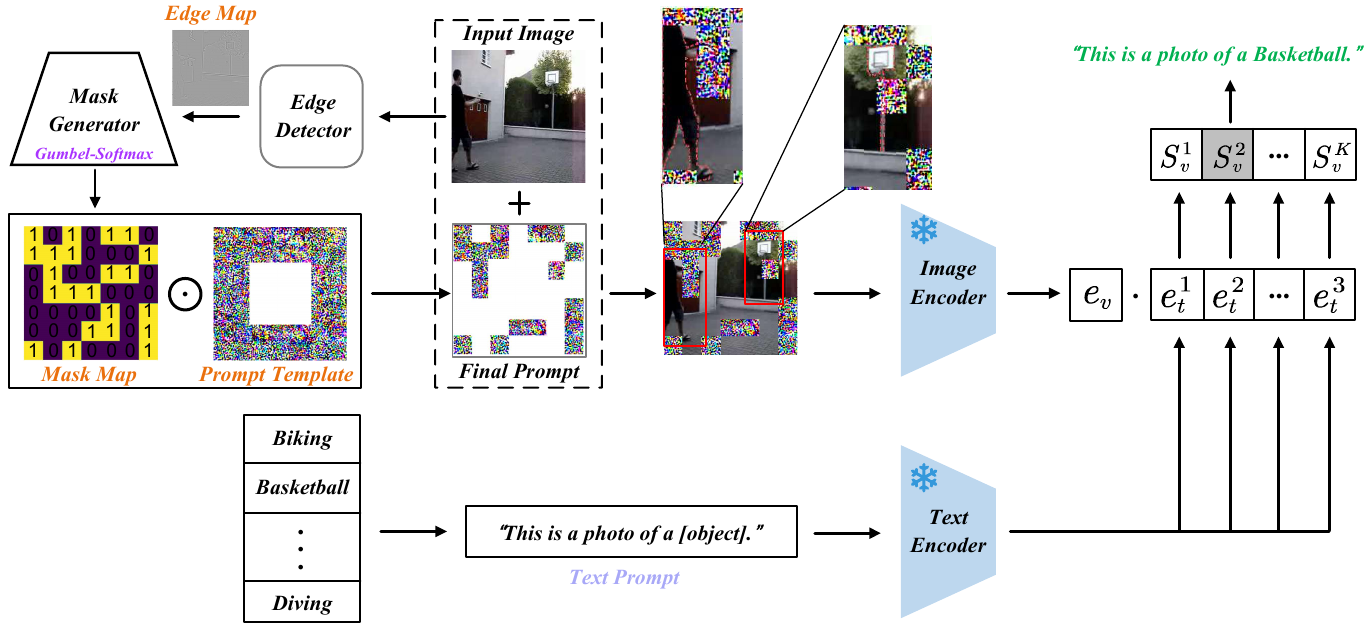}}
    \vspace{-0.1 in}
    \caption
    { 
   The overall architecture of our AdaViPro, which mainly consists of an edge detector and a mask generator. 
   We retain the original pipeline of VP as stated in Section \ref{vp}.
   Preferably viewed in color.
    }
    \vspace{-0.1 in}
    \label{mathod overview}
\end{figure*}

\section{Related Work}
\subsection{Large-scale Models}
With the advancement of self-supervised learning \cite{bert,mae} and the availability of large-scale datasets, the problem of `data hunger' in high-capacity deep learning models has been effectively addressed. 
In this context, the scalability potential of the transformer \cite{transformer} is continuously explored, leading to the development of large-scale models in NLP that exhibit formidable capabilities \cite{gpt3}.
This success has also been mirrored in the field of computer vision \cite{mae} and multi-modal learning \cite{clip,liang2023language}.
Recent works propose to pre-train large-scale foundational visual models via the self-supervised learning pipeline, including methods like contrastive learning \cite{clip} and masked image modeling \cite{mae}, among others.
Our work is orthogonal to the pretraining of foundational models. 
We focus on how to transfer the generalized representations of large pre-trained models to downstream tasks.


\subsection{Parameter-efficient Fine-tuning}
With the increasing scale of training data and model size, fully fine-tuning the entire model for each downstream task becomes prohibitively expensive and impractical in terms of training cost and model storage \cite{actionclip,dam-vp,liu2023cost}.
To remedy this, parameter-efficient fine-tuning (PEFT) is first introduced in natural language processing \cite{prompt,prefixtuning}.
Recently, prompt fine-tuning has also been studied in computer vision.
VPT \cite{vpt} first appends some learnable tokens to the tokenized sequence of the input image.
VP \cite{vp}, on the other hand, modifies images in the pixel space.
It superimposes extra learnable noises as task-specific `mosaics' onto the input image.
However, these methods effectively optimize `what to add', while overlooking the equally important aspect of `where to add', which is currently determined by hand-crafted strategy.
In contrast, our method jointly optimizes `what to add' and `where to add' within an end-to-end training pipeline, to adaptively generate image-specific visual prompts.

\section{Methods}
\subsection{Preliminaries}
\label{vp}
In this paper, we build AdaViPro based on the prompt method VP \cite{vp}, which is a kind of model-agnostic universal pixel-level prompt. 
Note that the techniques introduced in this paper are versatile and can also be applied to other prompt methods.
To provide legibility and context, we first review the main idea of VP.
Specifically, taking CLIP as an example, let $f\left( \cdot \mid \theta _v \right) $ and $f\left( \cdot \mid \theta _t \right) $ denote the vision and text encoder respectively.
Given the category collection $C = \{c_1, c_2,...,c_K\}$ and image-text pair $M = \{x_v, c_v\}$, VP can be formulated as:
\begin{equation}
    e_v = f\left( x_v \oplus  v_{\varphi} \mid \theta _v \right), e_{t}^{c} = f\left( c_i \mid \theta _t \right),
\end{equation}
where $v_{\varphi}$ is the pixel-level prompt of VP, which has the same size as the input image. 
$\oplus$ denotes the element-wise summation of matrices. 
Note that here we overlook the description of the text prompt in the equation, as it is not the focus of VP.

Following this, we can get the similarity score $S_{v}^{c}$ of image $x_v$ and category $e_{t}^{c}$ as:
\begin{equation}
    S_{v}^{c} = s(e_v, e_{t}^{c}),
\end{equation}
where the $s(\cdot\mid\cdot)$ denots the cosine similarity function.

For all categories in $C$, the text encoder pre-computes their embedding, and the input image embedding $e_v$ is computed for similarities with all the categories embeddings respectively.
During training, both the vision encoder $f\left( \cdot \mid \theta _v \right) $ and text encoder $f\left( \cdot \mid \theta _t \right) $ are all frozen.
The parameters of the prompt $v_{\varphi}$ are optimized by maximizing the similarity corresponding to the ground-truth category.

\subsection{Method Overview}
Figure \ref{mathod overview} illustrates an overview of our proposed AdaViPro, which mainly consists of an edge detector and a mask generator.
The primary motivation of AdaViPro is to optimize `where to add' of prompts, for generating image-specific prompts. 
We reconceptualize the `where to add' question as a decision-making process of whether to enable the prompt for each region.
However, for pixel-level prompts, deciding `where to add' is not feasible.
Due to the exponential increase in potential configurations as image size grows, treating the task of generating an optimal prompt as a pixel-level dense prediction task becomes untouchable.
Therefore, we divide the image into multiple rectangular regions, serving as the smallest decision-making unit, which is similar to the patch in ViT, such as $16\times16$ or $32\times32$.

During inference, given an image, the edge detector first generates rough object outlines of the image with an off-the-shelf edge detection algorithm.
Then the mask generator previews the edge feature map and generates a prompt mask map.
Each region of the image has a corresponding mark in the mask map, which is composed of $\{0,1\}$, indicating whether to use or discard the prompt in its area.
During training, Gumbel-Softmax sampling is leveraged to facilitate the learning of mask generator, preserving the end-to-end learning through standard back-propagation.

\subsection{Learning Region-based Adaptive Visual Prompt}
\textbf{\textit{Edge Detector.}}
As previously mentioned, our objective is to generate position-adaptive visual prompts that circumvent overwriting the objects within the input images.
With this in mind, to direct the mask generator's focus towards the size and distribution of objects within the image, we initially preprocess the image into an edge map.
This procedure streamlines the feature extraction process of the mask generator and ensures its lightweight nature.

To this end, here we adopt the off-the-shelf laplacian edge detection algorithm, incorporated within the edge detector.
Note that any edge detection algorithm can be adapted.
Given an input image $x_v$, for each pixel $p = f \left(i,j \right)$ in the image, the edge map $m_e$ can be computed by the sum of the second-order partial derivatives with respect to the pixel coordinates $i$ and $j$, which is thus formulated as:
\begin{equation}
    m_e\left(i,j \right) = \Delta f (i,j)= \frac{\partial^2 f (i,j)}{\partial i^2} + \frac{\partial^2 f (i,j)}{\partial j^2}.
\end{equation}

For parallel computation and implementation simplicity, we employ a fixedweight convolution kernel, an approximation of the Laplacian operator, as the edge detector.

\textbf{\textit{Mask Generator.}}
The generator mainly consists of a convergence module $F_c$ and a policy module $F_p$.
Both are made up of the basic convolutions.
Convergence module $F_c$ is responsible for observing the local edge information and realizing the information transfer and aggregation of all regions.
Following this, policy module $F_p$ standardizes the size of output map $m_p$, so as to match the preset decision region size, and then maps the logits to $\{0,1\}$ decision variables.

\begin{table*}[ht]
    \vspace{-0.2 in}
    \caption{
    Comparison with the baseline methods across 9 datasets with VP's \cite{vp} default prompt size 30. 
    Fully-FT, Linear-FT, TP, and VP denote fully fine-tuning, linear fine-tuning, text prompt, and visual prompt, respectively.
    }
    \label{Comparison with the baseline methods}
    \centering
    \setlength{\tabcolsep}{2pt}
    \renewcommand{\arraystretch}{1.3}
    \begin{adjustbox}{max width=0.7\textwidth}
        \begin{tabular}{c|ccccccccc|c}
          \toprule
           & CIFAR10 & CIFAR100 & DTD & Flowers & Food101 & SUN397 & EuroSAT & UCF101 & Pets & Average\\
           \midrule
           Fully-FT & 95.8 & 82.1 & 72.3 & 97.4 & 80.5 & 64 & 97.9 & 80.9 & 88.5 & 84.4\\
           Linear-FT & 95.0 & 80.0 & 74.6 & 96.9 & 84.6 & 75.0 & 95.3 & 83.3 & 89.2 & 86.0 \\
           \midrule
           TP& 89.0 & 63.1 & 43.0 & 61.9  & 79.8 & 60.0 & 40.0 & 59.9 & 85.9 & 64.7\\
           VP& 94.2 & 75.3 & 57.1 & 70.3 & 78.9 & 60.6 & 96.4 & 66.1 & 85 & 76.0\\  
           \cellcolor{gray!20}AdaViPro& \cellcolor{gray!20}95.4 & \cellcolor{gray!20}76.0 & \cellcolor{gray!20}63.8 & \cellcolor{gray!20}71.1 & \cellcolor{gray!20}80.3 & \cellcolor{gray!20}62.4 & \cellcolor{gray!20}96.8 & \cellcolor{gray!20}70.2 & \cellcolor{gray!20}87.5 & \cellcolor{gray!20}\textbf{78.2} \perplus{2.2\%}\\
          \bottomrule
        \end{tabular}
    \end{adjustbox}
     \vspace{-0.2 in}
\end{table*}

\begin{table*}[ht]
  \begin{center}
    \caption{Comparison with the baseline method VP \cite{vp} in various prompt sizes of \{5, 60, 90, 112\}.
    As prompt size increases, the superiority of the adaptive prompt is more pronounced.
    See Figure \ref{fig2} for visualization of the results.
    }
    \label{Comparison in various prompt sizes}
    \setlength{\tabcolsep}{4pt}
    \renewcommand{\arraystretch}{1.3}
    \begin{adjustbox}{max width=0.91\textwidth}
        \begin{tabular}{c|cccc|cccc|cccc|cccc|c}
          \toprule
            & \multicolumn{4}{c|}{CIFAR10} & \multicolumn{4}{c|}{CIFAR100} & \multicolumn{4}{c|}{DTD} & \multicolumn{4}{c|}{UCF101} & \multirow{2}*{Average}\\
           \cmidrule{2-17}
           Prompt width & 5 & 60 & 90 & 112 & 5& 60 & 90 & 112 & 5 & 60 & 90 & 112& 5 & 60 & 90 & 112 &\\
           \midrule
           VP & 92.1 & 90.7 & 89.9 & 89.6 & 69.5 & 66.1 & 65.4 & 65.8 & 47.4 & 44.4 & 38.9 & 38.4 & 65.5 & 59.2 & 55.2 & 54.5 & 64.5\\  
           \cellcolor{gray!20}AdaViPro & \cellcolor{gray!20}92.4 & \cellcolor{gray!20}94.6 & \cellcolor{gray!20}94.2 & \cellcolor{gray!20}93.8 & \cellcolor{gray!20}71.1 & \cellcolor{gray!20}71.9 & \cellcolor{gray!20}71.7 & \cellcolor{gray!20}71.8 & \cellcolor{gray!20}48.5 & \cellcolor{gray!20}62.7 & \cellcolor{gray!20}62.4 & \cellcolor{gray!20}63.4 & \cellcolor{gray!20}68.0 & \cellcolor{gray!20}70.1 & \cellcolor{gray!20}69.9 & \cellcolor{gray!20}69.1 & \cellcolor{gray!20}\textbf{73.5}  \perplus{9.0\%} \\
          \bottomrule
        \end{tabular}
    \end{adjustbox}
    \vspace{-0.3 in}
  \end{center}
\end{table*}

During inference, given the prompt template $P$ and edge feature map $m_e$, the mask map $m_p$ can be generated as:
\begin{equation}
    m_p = \varPhi \left(  F_p\left( F_c\left( m_e \right) \right) \right) ,
\end{equation}
where $\varPhi$ is the dilation function that restores the region-based mask to the original image size.
Following this, the mask is applied point-wise to the prompt template, which gives the final image-specific prompt $\hat{P}$:
\begin{equation}
    \hat{P}=P\odot m_p,
\end{equation}
where $\odot$ denotes the element-wise matrix multiplication.

\textbf{\textit{Training with Gumbel-Softmax Sampling.}}
AdaViPro makes discrete decisions, denoted by  $\{0,1\}$, to determine whether to enable prompts in different regions.
Although the decision process is straightforward, discrete decision generation renders the policy module of the mask generator non-differentiable. 
To avoid such a complex hybrid training pipeline, we introduce Gumbel-Softmax sampling to solve the non-differentiability for end-to-end optimization.

Gumbel-Softmax sampling allows for differentiable sampling from a discrete probability distribution, providing an alternative to approximate the gradients of discrete operations within the computation graph.
To be specific, during training, the policy module $F_p$ first normalizes the logit map from convergence module $F_c$, according to the predefined size of the region size.
Subsequently, the Gumbel-Max trick is employed to generate discrete decision marks $\hat{P}_{i,j}$ for each region as:
\begin{equation}
\hat{P}_{i,j}=\underset{k \in\{0,1\}}{\arg \max }\left(\log e_{k, (i,j)}+G_{k, (i,j)}\right),
\label{gumbel1}
\end{equation}
where $G_{k, (i,j)} = -\log(-\log U_{k, (i,j)})$ is the standard gumbel distribution and $U_{k, (i,j)}$ is sampled from the uniform distribution $Uniform(0,1)$.
Since the \textit{argmax} operation is non-indifferentiable, a \textit{softmax} oprator is used to generate $\hat{P}_{i,j}$ as a one-hot vector as:
\begin{equation}
\hat{P}_{l,(i,j)}=\frac{\exp \left(\left(\log e_{l, (i,j)}+G_{l, (i,j)}\right) / \tau\right)}{\sum_{k \in\{0,1\}} \exp \left(\left(\log e_{k, (i,j)}+G_{k, (i,j)}\right) / \tau\right)},
\label{gumbel2}
\end{equation}
where $j \in \{ 0,1\}$, and $\tau$ is a temperature parameter, controlling the shape of the probability distribution, and adjusting the balance between exploration and exploitation.
It cools down as the training progresses, which is multiplied by a decay factor $\gamma$ at each epoch. 
Note that the Gumbel-Softmax sampling is only used to approximate the gradient during training.
During inference, we directly utilize the \textit{argmax} operation to generate discrete decision marks of mask map $m_p$.

\begin{table}[h]
    \vspace{-0.1 in}
    \caption{Comparisons of tunable parameters number.}
    \label{Comparisons of tunable parameters number}
    \centering
    \begin{adjustbox}{max width=0.48\textwidth}
        \begin{tabular}{c|cccc}
            \toprule
              & Fully-Tuning & Linear & VP & AdaViPro \\
            \midrule
            Total params (M)& $\sim$85.90 & $\sim$0.08 & 0.07 & 0.20 \\
            \bottomrule
        \end{tabular}
    \end{adjustbox}
    \vspace{-0.2 in}
\end{table}

\section{Experiments}
\subsection{Setup}
\textbf{\textit{Datasets and evaluation metrics.}}
For the comparison experiments, we evaluate the performance of our AdaViPro using nine datasets, namely CIFAR10 \& CIFAR100 \cite{cifar}, DTD \cite{dtd}, Food \cite{food}, SUN \cite{sun}, Flowers \cite{flower}, UCF101 \cite{ucf101}, Eurosat \cite{eurosat}, and Pets \cite{pets}.
Out of these, CIFAR 10, CIFAR 100, DTD, and UCF101 are used for ablation studies.

\textbf{\textit{Implementation Details.}} 
For the edge detector, we encapsulate the 4-neighborhood Laplacian kernel in a layer of 2D-CNN with a padding of 1.
Within the mask generator, we simplify its network structure as much as possible.
We use a single convolution layer in both the convergence module $F_c$ and the policy module $F_p$, respectively.
Unless stated otherwise, we follow training details in VP \cite{vp}. 
The initial temperature $\tau$ of Gumbel-Softmax sampling is set to 5.
The learning rate of the mask generator is set to 1, while 40 for the prompt template.
All the experiments are conducted on one NVIDIA RTX 3090 GPU with batch-size of 256.

\subsection{Comparison with Baseline Methods}
\textbf{\textit{Main Results.}} 
In this section, we compare the proposed method with the above-mentioned baseline methods to demonstrate the effectiveness of adaptive prompts.
Experiments are performed with multiple prompt sizes from 5 to 112.
Note that the prompt size refers to the width of a single-side prompt, which has four sides: top, bottom, left, and right. 
When the size is set to 112, the whole image will be completely covered by the prompt vector.

As indicated in Table \ref{Comparison with the baseline methods}, we first compare the performance of AdaViPro with several baselines across 9 datasets, utilizing the VP's \cite{vp} default prompt size 30.
The results demonstrate that AdaViPro outperforms TP and VP across all datasets by a considerable margin, with an average improvement of 2.2\%. 
Specifically, in the CIFAR10 and EuroSAT, AdaViPro exceeded the performance of Linear fine-tuning,  achieving impressive accuracy rates of 95.4\% and 96.8\%, respectively.
By strategically incorporating `where to add' into optimization, AdaViPro generates adaptive prompts that avoid obscuring objects, enhancing the recognition performance.


Table \ref{Comparison in various prompt sizes} shows the comparison results with various prompt sizes of \{5, 60, 90, 112\}.
AdaViPro outperforms VP for all prompt sizes across all datasets, with an average benefit of 9.0\%.
As prompt size increases, the performance of VP deteriorates, particularly in the DTD and UCF101.
\begin{figure*}[t]
    \vspace{-0.2 in}
    \centerline{\includegraphics[width=0.95\textwidth]{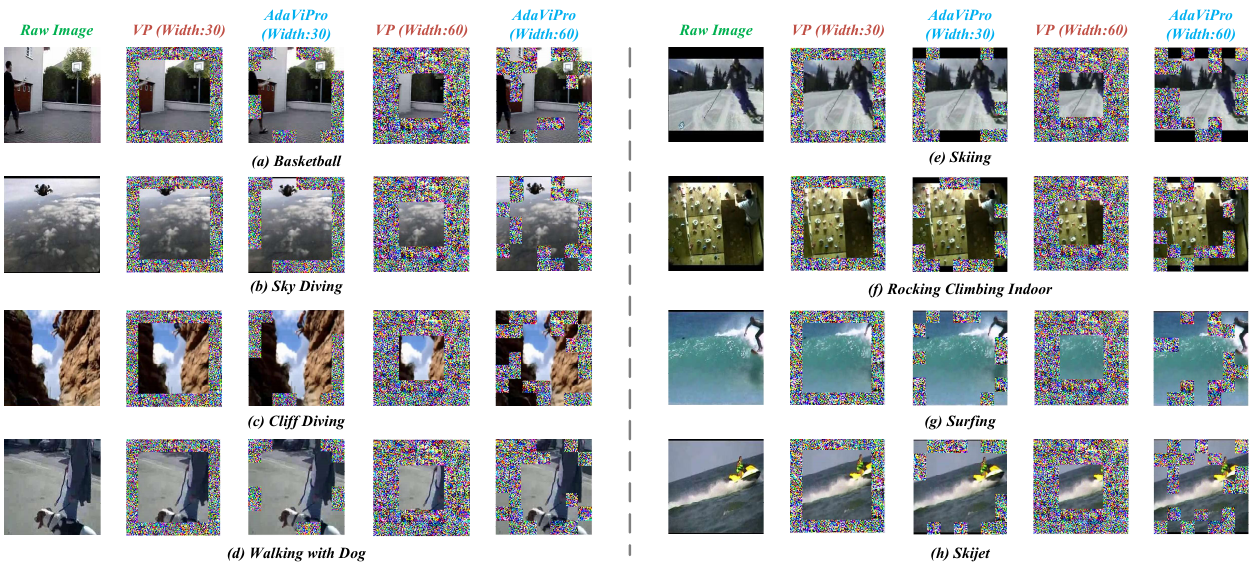}}
    \vspace{-0.1 in}
    \caption
    {   
       Qualitative examples showing the effectiveness of AdaViPro.
       1st column: Raw images serve as references for comparison.
       2nd and 4th column: VP with 30 and 60 widths.
       3rd and 5th column: Our AdaViPro with 30 and 60 widths.
    }
    \label{qualitative analysis}
    \vspace{-0.2 in}
\end{figure*}
With the prompt size of 112, the accuracy rates notably dropped to 38.4\% and 54.5\% in DTD and UCF101, respectively.
Similar results can also be observed in Figure \ref{fig2} as well.
VP’s performance undergoes a steep decline with the increase in prompt size.
Larger prompt sizes and an increased number of tunable parameters do not lead to performance improvements; rather, they result in the loss of key information in the input images.
In contrast, AdaViPro consistently maintains robust performance, even achieving the best results in DTD with a prompt size of 112.
This demonstrates that adaptive prompts enable instance-specific generation, effectively avoiding the obstruction of objects and information within the image.
	 	 	 	 	 	 	 	 
\textbf{\textit{Number of Tunable Parameters.}}
We compare the tunable parameters number of AdaViPro with different baseline methods, as shown in Table \ref{Comparisons of tunable parameters number}.
Fully fine-tuning requires adjusting all parameters of the model, which is quite expensive. 
The number of tunable parameters for linear fine-tuning and VP is roughly the same, but VP still exhibits a significant performance gap compared to the former.
Our AdaViPro adds only 0.13M additional parameters, enabling the generation of adaptive prompts, which achieves a well-balanced trade-off between accuracy and parameter efficiency.

\begin{table}[ht]
    \vspace{-0.1 in}
    \caption{Ablations on edge detection. w/ and w/o denote `with' or `without'. The default setting is marked in \colorbox{defaultcolor}{grey}.}
    \label{Ablations on edge detection}
    \centering
    \begin{adjustbox}{max width=0.42\textwidth}
        \begin{tabular}{c|cccc}
            \toprule
            Edge Detection & CIFAR10 & CIFAR100 & DTD & UCF101  \\
            \midrule
            \cellcolor{gray!20}w/ &  \cellcolor{gray!20}95.4 & \cellcolor{gray!20}76 &  \cellcolor{gray!20}63.8 & \cellcolor{gray!20}70.2 \\
            w/o & 94.4 & 74.4 & 51.5 & 68.5 \\
            \bottomrule
        \end{tabular}
    \end{adjustbox}
    \vspace{-0.2 in}
\end{table}

\subsection{Ablation Studies}
\textbf{\textit{Edge Detection.}}
AdaViPro is designed to generate image-specific prompts adaptively, to protect the objects within the input image.
Therefore, to direct the mask generator's focus towards the contour of the objects, we first preprocess the edge map of the input image via the edge detector $F_e$.
Table \ref{Ablations on edge detection} studies whether the edge detection works.
Our ablation study reveals that the model with edge features has a significant advantage in performance across all four datasets, especially in DTD.
The model without edge features collapses with a dramatic decline of 12.3\%, in line with our initial hypothesis. 

\begin{table}[ht]
    \vspace{-0.2 in}
    \caption{Ablations on temperature decay factor $\gamma$.}
    \label{Ablations on temperature decay factor}
    \centering
    \begin{adjustbox}{max width=0.42\textwidth}
        \begin{tabular}{c|cccc}
            \toprule
            Decay Factor $\gamma$ & CIFAR10 & CIFAR100 & DTD & UCF101  \\
            \midrule
            0.99 &  94.8 & 75.7 & 61.5 & 70.0  \\
            \cellcolor{gray!20}0.98 & \cellcolor{gray!20}95.4 & \cellcolor{gray!20}76.0 & \cellcolor{gray!20}63.8 & \cellcolor{gray!20}70.2  \\
            0.94 & 94.7 & 75.4 & 60.2 & 69.2 \\
            0.90 & 94.7 & 75.1 & 61.3 & 69.5 \\
            \bottomrule
        \end{tabular}
    \end{adjustbox}
    \vspace{-0.1 in}
\end{table}

\textbf{\textit{Temperature of Gumbel-Softmax.}}
The temperature of Gumbel-Softmax sampling controls the sharpness of the probability distribution. 
A lower temperature results in a significantly higher probability for a particular class compared to others, whereas a higher temperature leads to a more uniform probability distribution. 
Following this, during backpropagation, adjusting the temperature affects the smoothness of the gradients. 
Consequently, modulating the temperature can assist the model in maintaining a balance between exploration and exploitation during the training process, avoiding local minima during the early training stages.
We ablate the temperature decay factor $\gamma$ in \{0.9, 0.94, 0.98, 0.99\}.
As shown in Table \ref{Ablations on temperature decay factor}, both excessively large and small decay factors are detrimental to learning. 
We set 0.98 as the default factor.

\begin{table}[h]
    \vspace{-0.2 in}
    \caption{Ablations on the size of mask region.}
    \label{Ablations on the size of mask region}
    \centering
    \begin{adjustbox}{max width=0.42\textwidth}
        \begin{tabular}{c|cccc}
            \toprule
            Region size & CIFAR10 & CIFAR100 & DTD & UCF101  \\
            \midrule
            8 &  92.6 & 69.1 & 55.8 & 64.2  \\
            16 & 92.3 & 69.4 & 57.4 & 66.7 \\
            \cellcolor{gray!20}32 & \cellcolor{gray!20}95.4 & \cellcolor{gray!20}76.0 & \cellcolor{gray!20}63.8 & \cellcolor{gray!20}70.2 \\
            56 & 94.4 & 73.2 & 59.6 & 70.9 \\
            \bottomrule
        \end{tabular}
    \end{adjustbox}
    \vspace{-0.15 in}
\end{table}

\textbf{\textit{Size of Mask Region.}}
To avoid pixel-level dense prediction, we have divided the image into multiple rectangular regions.
These regions serve as atomic decision units, facilitating the regionalized adaptive prompts.
Therefore, the region size influences the granularity of the adaptive generation.
As shown in Table \ref{Ablations on the size of mask region}, we experiment with various region sizes of \{ 8, 16, 32, 56\}.
Interestingly, on UCF101, the largest region size of 56 exhibits the best performance of 70.9\%, while others at the size of 32.
Considering the optimal results for all datasets, we use the size of 32 as the default.

\begin{table}[h]
    \vspace{-0.2 in}
    \caption{Ablations on the embedding dim of mask generator.}
    \label{Ablations on the embedding dim of mask generator}
    \centering
    \begin{adjustbox}{max width=0.48\textwidth}
        \begin{tabular}{c|cccc|c}
            \toprule
            Embedding Dim & CIFAR10 & CIFAR100 & DTD & UCF101 & Parameter  \\
            \midrule
            8 &  94.1 & 74.8 & 61.6 & 68.3 & 0.09M  \\
            16 & 94.5 & 74.9 & 62.5 & 68.9 & 0.10M \\
            32 & 94.6 & 94.7 & 62.9 & 69.1 & 0.14M  \\
            \cellcolor{gray!20}64 & \cellcolor{gray!20}95.4 & \cellcolor{gray!20}76.0 & \cellcolor{gray!20}63.8 & \cellcolor{gray!20}70.2 & \cellcolor{gray!20}0.20M \\
            \bottomrule
        \end{tabular}
    \end{adjustbox}
    \vspace{-0.1 in}
\end{table}

\textbf{\textit{Embedding Dimension of Mask Generator.}}
As shown in Table \ref{Ablations on the embedding dim of mask generator}, we conducted an ablation study on the embedding dimension of the mask generator.
It affects the effectiveness of convergence and also influences the number of tunable parameters of the model.
When the embedding dimension is set to 8, the number of tunable parameters is only 0.09M, which is almost on par with the original 0.07M of VP.
Given that the increase in parameters is within an acceptable range, we greedily use the best-performing 64-dimension as the default.

\section{Qualitative Results}
Figure \ref{qualitative analysis} shows the comparison results of VP \cite{vp} and AdaViPro across various categories.
We compare the prompt effect at two different sizes, 30 and 60, respectively.
AdaViPro consistently preserves the original information of the image across various prompt sizes.
However, due to the fixed prompts, VP cannot perceive the distribution of objects within the image, resulting in significant damage to key information directly related to the label.
With the prompt size of 60, in almost all the examples, key information directly related to the labels has been compromised, and even for us humans, it becomes nearly impossible to recognize these categories.

\section{Conclusion}
In this paper, we propose a region-based adaptive visual prompt method, AdaViPro, which incorporates the `where to add' optimization of the prompt into the learning process.
For end-to-end optimization, Gumbel-Softmax sampling is leveraged to resolve nondifferentiability due to discrete decision-making during training.
Extensive experiments on nine image benchmarks demonstrate that our method significantly enhances the effectiveness in adapting pre-trained models.
In the future, we plan to continue our research on achieving more fine-grained adaptive prompt generation. 

\bibliographystyle{IEEEtran}
\bibliography{main}

\begin{thebibliography}{10}
\providecommand{\url}[1]{#1}
\csname url@samestyle\endcsname
\providecommand{\newblock}{\relax}
\providecommand{\bibinfo}[2]{#2}
\providecommand{\BIBentrySTDinterwordspacing}{\spaceskip=0pt\relax}
\providecommand{\BIBentryALTinterwordstretchfactor}{4}
\providecommand{\BIBentryALTinterwordspacing}{\spaceskip=\fontdimen2\font plus
\BIBentryALTinterwordstretchfactor\fontdimen3\font minus \fontdimen4\font\relax}
\providecommand{\BIBforeignlanguage}[2]{{%
\expandafter\ifx\csname l@#1\endcsname\relax
\typeout{** WARNING: IEEEtran.bst: No hyphenation pattern has been}%
\typeout{** loaded for the language `#1'. Using the pattern for}%
\typeout{** the default language instead.}%
\else
\language=\csname l@#1\endcsname
\fi
#2}}
\providecommand{\BIBdecl}{\relax}
\BIBdecl

\bibitem{transformer}
A.~Vaswani, N.~Shazeer, N.~Parmar, J.~Uszkoreit, L.~Jones, A.~N. Gomez, {\L}.~Kaiser, and I.~Polosukhin, ``Attention is all you need,'' \emph{NIPS}, vol.~30, 2017.

\bibitem{bert}
J.~Devlin, M.-W. Chang, K.~Lee, and K.~Toutanova, ``Bert: Pre-training of deep bidirectional transformers for language understanding,'' \emph{arXiv preprint arXiv:1810.04805}, 2018.

\bibitem{gpt3}
T.~Brown, B.~Mann, N.~Ryder, M.~Subbiah, J.~D. Kaplan, P.~Dhariwal, A.~Neelakantan, P.~Shyam, G.~Sastry, A.~Askell \emph{et~al.}, ``Language models are few-shot learners,'' \emph{NIPS}, vol.~33, pp. 1877--1901, 2020.

\bibitem{vit}
A.~Dosovitskiy, L.~Beyer, A.~Kolesnikov, D.~Weissenborn, X.~Zhai, T.~Unterthiner, M.~Dehghani, M.~Minderer, G.~Heigold, S.~Gelly \emph{et~al.}, ``An image is worth 16x16 words: Transformers for image recognition at scale,'' \emph{arXiv preprint arXiv:2010.11929}, 2020.

\bibitem{wang2023improving}
J.~Wang, C.~Zhang, J.~Huang, B.~Ren, and Z.~Deng, ``Improving scene graph generation with superpixel-based interaction learning,'' in \emph{ACM MM}, 2023, pp. 1809--1820.

\bibitem{vimo}
Y.~Tian, M.~Yang, L.~Zhang, Z.~Zhang, Y.~Liu, X.~Xie, X.~Que, and W.~Wang, ``View while moving: Efficient video recognition in long-untrimmed videos,'' in \emph{ACM MM}, 2023, pp. 173--183.

\bibitem{liu2023fine}
W.~Liu, T.~He, C.~Gong, N.~Zhang, H.~Yang, and J.~Yan, ``Fine-grained music plagiarism detection: Revealing plagiarists through bipartite graph matching and a comprehensive large-scale dataset,'' in \emph{ACM MM}, 2023, pp. 8839--8848.

\bibitem{dam-vp}
Q.~Huang, X.~Dong, D.~Chen, W.~Zhang, F.~Wang, G.~Hua, and N.~Yu, ``Diversity-aware meta visual prompting,'' in \emph{CVPR}, 2023, pp. 10\,878--10\,887.

\bibitem{actionclip}
M.~Wang, J.~Xing, J.~Mei, Y.~Liu, and Y.~Jiang, ``Actionclip: Adapting language-image pretrained models for video action recognition,'' \emph{TNNLS}, 2023.

\bibitem{vp}
H.~Bahng, A.~Jahanian, S.~Sankaranarayanan, and P.~Isola, ``Exploring visual prompts for adapting large-scale models,'' \emph{arXiv preprint arXiv:2203.17274}, 2022.

\bibitem{coop}
K.~Zhou, J.~Yang, C.~C. Loy, and Z.~Liu, ``Learning to prompt for vision-language models,'' \emph{IJCV}, vol. 130, no.~9, pp. 2337--2348, 2022.

\bibitem{vpt}
M.~Jia, L.~Tang, B.-C. Chen, C.~Cardie, S.~Belongie, B.~Hariharan, and S.-N. Lim, ``Visual prompt tuning,'' in \emph{ECCV}.\hskip 1em plus 0.5em minus 0.4em\relax Springer, 2022, pp. 709--727.

\bibitem{ju2022prompting}
C.~Ju, T.~Han, K.~Zheng, Y.~Zhang, and W.~Xie, ``Prompting visual-language models for efficient video understanding,'' in \emph{ECCV}.\hskip 1em plus 0.5em minus 0.4em\relax Springer, 2022, pp. 105--124.

\bibitem{prompt}
T.~Gao, A.~Fisch, and D.~Chen, ``Making pre-trained language models better few-shot learners,'' \emph{arXiv preprint arXiv:2012.15723}, 2020.

\bibitem{moco}
K.~He, H.~Fan, Y.~Wu, S.~Xie, and R.~Girshick, ``Momentum contrast for unsupervised visual representation learning,'' in \emph{CVPR}, 2020, pp. 9729--9738.

\bibitem{mae}
K.~He, X.~Chen, S.~Xie, Y.~Li, P.~Doll{\'a}r, and R.~Girshick, ``Masked autoencoders are scalable vision learners,'' in \emph{CVPR}, 2022, pp. 16\,000--16\,009.

\bibitem{clip}
A.~Radford, J.~W. Kim, C.~Hallacy, A.~Ramesh, G.~Goh, S.~Agarwal, G.~Sastry, A.~Askell, P.~Mishkin, J.~Clark \emph{et~al.}, ``Learning transferable visual models from natural language supervision,'' in \emph{ICML}.\hskip 1em plus 0.5em minus 0.4em\relax PMLR, 2021, pp. 8748--8763.

\bibitem{liang2023language}
X.~Liang, D.~Wang, Q.~Wang, B.~Wan, L.~An, and L.~He, ``Language-guided visual aggregation network for video question answering,'' in \emph{ACM MM}, 2023, pp. 5195--5203.

\bibitem{liu2023cost}
Y.~Liu, M.~Yang, Y.~Tian, L.~Zhang, X.~Que, and W.~Wang, ``Cost-effective modality selection for video popularity prediction,'' in \emph{IJCNN}.\hskip 1em plus 0.5em minus 0.4em\relax IEEE, 2023, pp. 1--8.

\bibitem{prefixtuning}
X.~L. Li and P.~Liang, ``Prefix-tuning: Optimizing continuous prompts for generation,'' \emph{arXiv preprint arXiv:2101.00190}, 2021.

\bibitem{cifar}
A.~Krizhevsky, G.~Hinton \emph{et~al.}, ``Learning multiple layers of features from tiny images,'' 2009.

\bibitem{dtd}
M.~Cimpoi, S.~Maji, I.~Kokkinos, S.~Mohamed, and A.~Vedaldi, ``Describing textures in the wild,'' in \emph{CVPR}, 2014, pp. 3606--3613.

\bibitem{food}
L.~Bossard, M.~Guillaumin, and L.~Van~Gool, ``Food-101--mining discriminative components with random forests,'' in \emph{ECCV}.\hskip 1em plus 0.5em minus 0.4em\relax Springer, 2014, pp. 446--461.

\bibitem{sun}
J.~Xiao, J.~Hays, K.~A. Ehinger, A.~Oliva, and A.~Torralba, ``Sun database: Large-scale scene recognition from abbey to zoo,'' in \emph{CVPR}.\hskip 1em plus 0.5em minus 0.4em\relax IEEE, 2010, pp. 3485--3492.

\bibitem{flower}
M.-E. Nilsback and A.~Zisserman, ``Automated flower classification over a large number of classes,'' in \emph{ICVGIP}.\hskip 1em plus 0.5em minus 0.4em\relax IEEE, 2008, pp. 722--729.

\bibitem{ucf101}
K.~Soomro, A.~R. Zamir, and M.~Shah, ``Ucf101: A dataset of 101 human actions classes from videos in the wild,'' \emph{arXiv preprint arXiv:1212.0402}, 2012.

\bibitem{eurosat}
P.~Helber, B.~Bischke, A.~Dengel, and D.~Borth, ``Eurosat: A novel dataset and deep learning benchmark for land use and land cover classification,'' \emph{JSTARS}, vol.~12, no.~7, pp. 2217--2226, 2019.

\bibitem{pets}
O.~M. Parkhi, A.~Vedaldi, A.~Zisserman, and C.~Jawahar, ``Cats and dogs,'' in \emph{CVPR}.\hskip 1em plus 0.5em minus 0.4em\relax IEEE, 2012, pp. 3498--3505.

\end{thebibliography}

\end{document}